\documentclass[letterpaper, 10 pt, conference]{IEEEtran}  

\IEEEoverridecommandlockouts                              

\usepackage{graphics} 
\usepackage{epsfig} 
\usepackage{mathptmx} 
\usepackage{times} 
\usepackage{amsmath} 
\usepackage{amssymb}  
\usepackage{enumerate} 
\usepackage{subfig}
\usepackage{multirow}
\usepackage{mathtools}
\usepackage[english, onelanguage,ruled,vlined, longend, titlenotnumbered]{algorithm2e}

\newtheorem{definition}{Definition}
\newtheorem{theorem}{Theorem}

\newtheorem{rem}{Remark}

\newcommand{\RR}{{\mathbb R}}

\DeclareMathOperator*{\argmin}{argmin}
\DeclareMathOperator*{\argmax}{argmax}

\usepackage{siunitx}
\usepackage{pgfplots,pgfplotstable}
\usepackage{tikz}
\usetikzlibrary[patterns]
\pgfplotsset{compat=1.12}
\usetikzlibrary{plotmarks}
\usepackage[nospace]{cite}
\usetikzlibrary{arrows,shapes,positioning}
\usetikzlibrary{calc,trees,positioning,arrows,chains,shapes.geometric,%
	decorations.pathreplacing,decorations.pathmorphing,shapes,%
	matrix,shapes.symbols}
\usetikzlibrary{decorations.markings}

\usepackage[top=54pt, bottom=54pt, left=54pt,
      right=54pt]{geometry}

\usepackage{setspace}

\usepackage{hyperref}
\hypersetup{hidelinks}

\author{P. Chauchat, A. Barrau, S. Bonnabel
 \thanks{Paul Chauchat and Silvere Bonnabel are with MINES ParisTech, PSL Reasearch University, Centre for Robotics, 60 bd Saint-Michel, 75006 Paris, France
         {\tt\small paul.chauchat@mines-paristech.fr, silvere.bonnabel@mines-paristech.fr}}%
 \thanks{Axel Barrau is with SAFRAN TECH, Groupe Safran, Rue des Jeunes Bois - Ch\^ateaufort, 78772 Magny Les Hameaux CEDEX, France
         {\tt\small axel.barrau@safrangroup.com}}%
}

\begin{document}
\title{\vspace*{3.5mm} Invariant smoothing on Lie Groups}
\author{\IEEEauthorblockN{Paul \textsc{Chauchat},
		Axel  \textsc{Barrau} and
		Silv\`ere \textsc{Bonnabel}}%
    \thanks{Paul Chauchat and Silvere Bonnabel are with MINES ParisTech, PSL Reasearch University, Centre for Robotics, 60 bd Saint-Michel, 75006 Paris, France
         {\tt\small paul.chauchat@mines-paristech.fr, silvere.bonnabel@mines-paristech.fr}}%
 \thanks{Axel Barrau is with SAFRAN TECH, Groupe Safran, Rue des Jeunes Bois - Ch\^ateaufort, 78772 Magny Les Hameaux CEDEX, France
         {\tt\small axel.barrau@safrangroup.com}}%
    }

\maketitle
\begin{abstract}
In this paper we propose a (non-linear) smoothing algorithm for group-affine observation systems, a recently introduced class of estimation problems on Lie groups that bear a particular structure. As most non-linear smoothing methods, the proposed algorithm  is based on a maximum a posteriori estimator, determined by optimization. But owing to the specific properties of the considered class of problems, the involved linearizations are proved to have a form of independence with respect to the current estimates, leveraged to avoid (partially or sometimes totally) the need to relinearize. The method is validated on a robot localization example, both in simulations and on real experimental data.  
\end{abstract}
\section{INTRODUCTION}

Statistical state estimation based on sensor fusion has played a major role in localization for the last decades. Most of the proposed solutions revolved around filtering, especially under the framework of the standard Extended Kalman Filter (EKF) \cite{farrell2008aided}. Many alternative filters have been introduced to overcome the inherent limitations of the EKF, such as particle filters or the Unscented Kalman Filter. In the robotics field, and especially in the Simultaneous Localization and Mapping (SLAM) community, the smoothing approach, based on the Gauss-Newton (GN) algorithm, was proposed in order to take advantage of the sparsity of the involved matrix, and reduce the consequences of wrong linearization points \cite{dellaert2006square}. It had a tremendous impact and led to many of the state-of-the-art algorithms for SLAM and visual odometry systems \cite{kaess2012iSAM2,forster2017SVO2,rosen2017SE-sync}, to cite a few. It was more recently applied to GPS aided inertial navigation, showing promising results \cite{zhao2014differential,indelman2012factor}.
In the meantime, the advantages of the Lie group structure for probabilistic robotics, and especially that of the configuration spaces $SE(d)$, have been widely recognized \cite{chirikjian2009stochastic1,chirikjian2011stochastic2,barfoot2014associating,wolfe2011bayesian,diemer2015controller,hertzberg2013integrating}. This led to new non-linear filtering algorithms \cite{bonnabel2008symmetry,bourmaud2015concentrated,barrau2014invariant} which showed remarkable properties \cite{barrau2014invariant,barrau2015SLAM}, and to geometrical methods at the core of all the recent solvers for smoothing problems involving relative poses \cite{rosen2017SE-sync,forster2017SVO2}.

The smoothing technique proposed in this paper builds upon the invariant filtering framework, whose main results can be found in \cite{bonnabel2008symmetry,barrau2014invariant}. An overview of the theory and its applications is made in \cite{barrau2017annual}, and a focus on the case of SLAM in \cite{barrau2015SLAM}. Similar results were achieved with observers \cite{mahony2017geometric}. A more tutorial approach to the theory, along with some new results, can also be found in the recent paper \cite{barrau2017linear} dedicated to the discrete-time case.
In the present paper, we devise a smoothing framework adapted to robot localization, where the dependence of the whole information matrix of the system with respect to the linearization points is reduced, or even removed under mild conditions. This brings immediate benefits: 1) The number of iterations to convergence is reduced 2) The consistency of the smoother is improved.
As usual with estimation methods derived from the theory of invariant filtering this comes at a price: the framework is adapted to a specific class of systems called \emph{group-affine observation systems} defined in \cite{barrau2014invariant} (the equations of inertial navigation \cite{barrau2014invariant}, as well as SLAM \cite{barrau2015SLAM}, fall within this framework).

\subsection{Links and differences with the previous literature}

Smoothing on manifolds is well established, and a thorough study of a Gauss-Newton algorithm on Riemannian manifolds can be found in \cite{absil2008optim}. A lot of work has been dedicated to studying the non-Euclidean structure of the state-space of SLAM \cite{chirikjian2009stochastic1,chirikjian2011stochastic2,barfoot2017robotics}. Kuemmerle et al. introduced pose-graph algorithms for general manifolds \cite{grisetti2010hierarchical,kuemmerle2011g2o}. Most of the recent work on SLAM focused on the 2D and 3D Special Euclidean groups $SE(2)$ and $SE(3)$ which are used to represent position-orientation states, especially in the pose-SLAM framework, or on products $SE(3)\times \RR^n$. The majority uses the group structure to define the cost function, and we then recover a group synchronization problem \cite{boumal2013rotations}. To solve it, most methods revolve around using the manifold structure of the state-space and custom retractions \cite{forster2016preintegration} or using the matrix structure to relax the problem \cite{martinec2007robust,rosen2015relaxation,rosen2017SE-sync}.
The recent work of Bourmaud et al \cite{bourmaud2016iterated,bourmaud2016online} is, to the authors' knowledge, the only one which fully relies on the Lie group structure for the estimate. In a way, the proposed method can be seen as an instance of this algorithm, but where the choice of the uncertainty representation is commanded by the system instead of being free, and an additional trick regarding the observation factors provides new properties.

These approaches are focused on leveraging the intrinsic tools of a given structure of the state space (Riemannian manifold or Lie group). For instance in \cite{bourmaud2016iterated}, the proposed algorithm only required the state to belong to a Lie group, but no assumption is made regarding the dynamics and observation functions. In the present paper, the equations of the system are assumed to have specific properties with respect to the chosen group structure. This reduces the application field, but ensures in turn striking properties.

\subsection{Organization of the paper}
The paper is divided as follows. Section \ref{sec:lie_groups} contains mathematical preliminaries. Section \ref{sec:classic_smoothing} recalls the standard framework of smoothing, both in the vector and manifold settings. Section \ref{sec:smoothing_group} presents our smoothing algorithm for   group-affine systems on Lie groups, which we call invariant smoothing, and which is the main contribution of the paper. Section \ref{sec:exp} illustrates the theory with both a simulated and an experimental 2D robot localization problems, showing the advantage of the proposed method over classical smoothing algorithms, and also over the recently developed (invariant) Kalman filter based  UKF-LG  and IEKF (Invariant Extended Kalman Filter)\cite{brossard2017unscented,barrau2014invariant}.

\section{Mathematical preliminaries}\label{sec:lie_groups}

\subsection{Lie Groups}
In this section we recall the definitions and basic properties of matrix Lie groups, Lie algebra and random variables on Lie groups. A matrix Lie group $G \subset \RR^{N\times N}$ is a set of square invertible matrices that is a group, i.e., the following properties hold:
\begin{equation}
 I_N \in G;~ \forall \chi \in G, \chi^{-1} \in G;~ \forall \chi_1, \chi_2 \in G, \chi_1 \chi_2 \in G
\end{equation}
Locally about the identity matrix $I_N$, the group $G$ can be identified with an vector space $\RR^q$ using the matrix exponential map $\exp_m(.)$, where $q = \dim G$. Indeed, to any $\xi \in \RR^q$ one can associate a matrix $\xi^{\wedge}$ of the tangent space of $G$ at $I_N$, called the Lie algebra $\mathfrak{g}$. We then define the exponential map $\exp : \RR^q \rightarrow G$ for Lie groups as
\begin{align}\label{expmap}
 \exp\left(\xi\right) = \exp_m\left(\xi^{\wedge}\right),
\end{align}\emph{Locally}, it is a bijection, and one can define the Lie logarithm map $\log : G \rightarrow \RR^q$ as its inverse:
$
  \log\left(\exp\left(\xi\right) \right)= \xi.
$

As Lie groups are not necessarily commutative, in general we have $\exp(x) \exp(y) \neq \exp(x + y)$ for $x,y \in \mathfrak{g}$. However, the vector $z$ such that $\exp(x)\exp(y) = \exp(z)$ satisfies the Baker-Campbell-Hausdorff (BCH) formula \cite{gallier2016diffgeo} $z = BCH(x,y) = x + y + r(x,y)$, where $r(x,y)$ contains higher order terms.

A last classical tool is the (inner) automorphism $\Psi_a \in Aut(G)$ defined for each $a \in G$ as $\Psi_a : \quad g \mapsto a g a^{-1}$. Its differential at the identity element $Id$ of $G$ is called \emph{adjoint operator} and denoted by $Ad_a : \mathfrak{g} \mapsto \mathfrak{g}$. It satisfies : 
$$
\forall a \in G, u \in \mathfrak{g},\ a \exp(u) a^{-1} = \exp(Ad_a u).
$$

\vspace*{-0.1cm}
\subsection{Uncertainties on Lie Groups}
To define random variables on Lie groups, we cannot apply the usual approach of additive noise for $\chi_1, \chi_2 \in G$ as $G$ is not a vector space, i.e., generally $\chi_1 + \chi_2 \notin G$ does not hold. In contrast, we adopt the framework of  \cite{barfoot2014associating}, see also \cite{barrau2015intrinsic}, which is slightly different from the pioneering approach of \cite{chirikjian2009stochastic1,chirikjian2011stochastic2}.
Indeed, we define the probability distribution $\chi \sim \mathcal{N}_L(\bar{\chi},\mathbf{P})$ for the random variable $\chi \in G$ as 
\begin{equation}
 \chi =  \bar{\chi} \exp \left(\xi\right), \text{~} \xi \sim \mathcal{N}\left(\mathbf{0}, \mathbf{P}\right), \label{eq:left}
\end{equation}
where $\mathcal{N}\left(.,.\right)$ is the classical Gaussian distribution in Euclidean space and $\mathbf{P}\in \RR^{q \times q}$ is a covariance matrix. In the sequel, we will refer to \eqref{eq:left} as the left-invariant Gaussian distribution on $G$, owing to the fact that the discrepancy  $\bar{\chi}^{-1} \chi $ which is invariant to left multiplications $(\bar\chi,\chi)\mapsto(\Gamma\bar\chi,\Gamma\chi)$, is the exponential of a Gaussian $\xi$. In \eqref{eq:left}, the noise-free quantity $\bar{\chi}$ is viewed as the mean, and the dispersion arises  through left multiplication with the exponential of a Gaussian random variable. Similarly, the distribution $\chi \sim \mathcal{N}_R(\bar{\chi},\mathbf{P})$ can be defined through right multiplication  as
\begin{equation}
 \chi = \exp \left(\xi\right) \bar{\chi}, \text{~} \xi \sim \mathcal{N}\left(\mathbf{0}, \mathbf{P}\right). \label{eq:right}
\end{equation}
We stress that we have defined these probability density functions directly in the vector space $\RR^q$ such that both $\mathcal{N}_L\left(.,.\right)$ and $\mathcal{N}_R\left(.,.\right)$ are not Gaussian distributions. 


\section{Classical smoothing}
\label{sec:classic_smoothing}


\subsection{Maximum a posteriori (MAP) estimate}
Contrarily to filtering approaches, which only consider the latest state of the system, smoothing methods aim at recovering the maximum a posteriori estimate (MAP) of the full trajectory given a set of measurements $Z=(y_1,\cdots,y_m)$:
\begin{equation}
\chi_0^*, \cdots, \chi_n^* = \argmax_{(\chi_i)_{i \leq n}} p((\chi_i)_{i \leq n}\mid Z)
\end{equation}
Under a markovian assumption on the trajectory and conditional independence of the measurements, and given a prior on the initial state $\chi_0$, this becomes
\begin{equation}
\chi_0^*, \cdots, \chi_n^* = \argmax_{(\chi_i)_{i \leq n}} p(\chi_0)  \prod_i p(\chi_{i+1} | \chi_i) \prod_k p(y_k | (\chi_i)_{i\leq n} )
\label{eq:MAP}
\end{equation}
In this paper, we consider partial measurements of a single state, i.e. we have for some instant $t_k$:
\begin{equation}
\forall k,\ p(y_k | (\chi_i)_{i\leq n} ) = p(y_k | \chi_{t_k})
\end{equation}

\subsection{Gaussian Smoothing on vector spaces}
In the general case, \eqref{eq:MAP} represents a difficult high dimensional optimization problem. However, if the noises are supposed to be additive and Gaussian, this boils down to a non-linear least-squares problem \cite{barfoot2017robotics}.

Consider the following prior, dynamics, and measurement model:
\begin{equation}
\chi_0 = \bar{\chi}_0 + z_0 \qquad \chi_{i+1} = f_i(\chi_i) + w_i \qquad y_k = h(\chi_{t_k}) + v_k
\label{eq:noisy_linear_system}
\end{equation}
where $z_0$, $w_i$ and $v_k$ are gaussian white noises of respective covariances $P_0$, $Q_i$ and $N_k$.
Then the MAP estimates satisfies:
\begin{align}
\chi_0^*, \cdots, \chi_n^* = \argmin_{(\chi_i)_{i \leq n}} 
\|\chi_0 &- \bar{\chi}_0\|_{P_0}^2  
+ \sum_i \|f_i(\chi_i) - \chi_{i+1}\|_{Q_i}^2 \nonumber \\
&+ \sum_k \|h(\chi_{t_k}) - y_k\|_{N_k}^2
\label{eq:additive_MAP}
\end{align}
where we used the Mahalanobis norm $\|e\|_\Sigma^2 = e^T \Sigma^{-1} e$. Solving \eqref{eq:additive_MAP} is done through iterative algorithms involving sequential linearizations, such as Gauss-Newton (GN) \cite{dellaert2006square, zhao2014differential}.

For  $i \leq n$, denoting by $\hat{\chi}_i$ the current estimates and letting   $\delta \chi_i$ be the search parameter for the current iteration, each iteration consists in  solving the following least-squares problem, based on a first-order Taylor expansion of \eqref{eq:additive_MAP}
\begin{align}
(\delta \chi_i)_{i\leq n} = \argmin_{(\delta \chi_i)_{i \leq n}} 
\|p_0 &+ \delta \chi_0\|_{P_0}^2  
+ \sum_i \|a_i + F_i \delta \chi_i - \delta \chi_{i+1}\|_{Q_i}^2  \nonumber \\
&+ \sum_k \|c_k + H_k \delta \chi_{t_k}\|_{N_k}^2
\label{eq:linear_lq}
\end{align}
where $p_0 = \hat{\chi}_0 - \bar{\chi}_0$, $a_i = f_i(\hat{\chi}_i) - \hat{\chi}_{i+1}$, $c_k = h(\hat{\chi}_{t_k}) - y_k$, $F_i$ and $H_k$ are the Jacobians of $f_i$ and $h_{t_k}$ evaluated at $\hat{\chi}_i$ and $\hat{\chi}_{t_k}$ respectively.
Each of the terms of the sum are referred to as factors. This name comes from the link between the MAP estimation problem and graphical models, in particular factor graphs \cite{dellaert2006square}, which lead to powerful solvers \cite{kaess2012iSAM2,rosen2014RISE,kuemmerle2011g2o} still used in many recent applications \cite{forster2017SVO2}.

In this paper, we focus on the non-linear least-squares problem which is to be solved, and argue that a better parametrization taking into account the geometry of the system will lead to better convergence properties, regardless of the solver used.

\subsection{Smoothing on manifolds}
A large body of literature was dedicated to   smoothing of systems which do not live in vector spaces but on a manifold $\mathcal{M}$ \cite{forster2016preintegration,grisetti2010hierarchical}. The main idea is to replace the $-$ and $+$ operators of the vector space by adapted ones $\boxminus$, $\boxplus$, which highlight the fact that $\mathcal{M}$ is not a vector space and that linearization takes place on its tangent space $T \mathcal{M}$, where the $\delta \chi_i$ should live. Thus, these operators are defined such that $\boxminus : \mathcal{M} \times \mathcal{M} \rightarrow T \mathcal{M}$ and $\boxplus : \mathcal{M} \times T \mathcal{M} \rightarrow \mathcal{M}$.

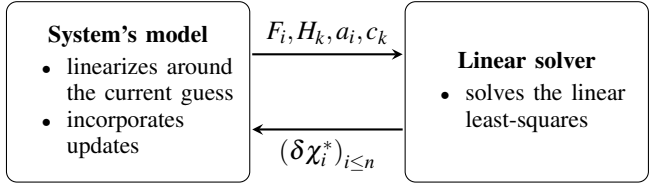
\begin{figure}[thpb]
       \centering
      \tikzstyle{monrectangle} = [rectangle, rounded corners, minimum width=3cm, minimum height=2.4cm,text centered, text width=3cm, draw=black]
\tikzstyle{arrow} = [thick,->,>=stealth]

\begin{tikzpicture}[node distance=5.3cm]
\node (system) [monrectangle] {\small \textbf{System's model} \begin{itemize}
	\item linearizes around the current guess
	\item incorporates updates
	\end{itemize} \par};
\node (linear) [monrectangle,right of=system] {\small \textbf{Linear solver} \begin{itemize}
	\item solves the linear least-squares
	\end{itemize} \par};
	\draw [arrow] (1.65,0.5)  -- node[above] {$F_i,H_k,a_i,c_k$} (3.65,0.5);
    \draw [arrow] (3.65,-0.5)  -- node[below] {$\left(\delta \chi_i^*\right)_{i\leq n}$} (1.65,-0.5);
\end{tikzpicture}
       \caption{Schematic representation of the two main components of a smoothing algorithm, the non linear model of the system which handles the transitions from the manifold to the tangent space and back, and the solver which inverts the obtained linear system.}
       \label{fig:schema}
    \end{figure}

Therefore, a smoothing algorithm can be divided into two main  parts, the non-linear and the linear ones. The former dictates how the jacobians and errors are computed, and how the updates are handled, while the latter is the solver,  only concerned with linear algebra, as illustrated on Figure \ref{fig:schema}. In this work, we focus on the modeling part, and propose a new framework which can be used with any existing solver.

When the manifold is  a Lie group, the group multiplication and its intrinsic operators $\log$ and $\exp$ naturally yield a family of such operators, through the following definitions:
\begin{align}
\boxplus_L (\chi, \xi) = \exp(\xi) \chi &\qquad \boxminus_L (\chi_1, \chi_2) = \log(\chi_1 \chi_2^{-1})\\
\boxplus_R (\chi, \xi) = \chi \exp(\xi) &\qquad \boxminus_R (\chi_1, \chi_2) = \log(\chi_2^{-1} \chi_1)
\end{align}
Note that, these operators were used in the definition of Section \ref{sec:lie_groups}. Smoothing methods on (Lie) groups were used extensively for the particular cases of pose-SLAM, visual odometry, and group synchronization problems in general \cite{rosen2017SE-sync,forster2017SVO2,boumal2013rotations}, which only consider relative measurements of the full states. They were recently extended to a more general case by Bourmaud et al \cite{bourmaud2016iterated}, but focusing on the Lie group structure essentially, without discussing the best choice of group structure with respect to the system's equations. In the next section, we focus our study on a particularly rich class of models on Lie groups.

\section{Invariant smoothing}
\label{sec:smoothing_group}


In this section, we consider smoothing on Lie groups for a special class of systems. It builds on the invariant filtering framework, hence its name. The remarkable result is that the dependency of the information matrix of smoothing  algorithms (partially) disappears, owing to the interplay between the group structure and the system's properties. 

\subsection{Group-affine observation systems}
In this work we argue that, just as  vector spaces are particularly well-suited to linear systems, the Lie group formalism is beneficial to a particular class of dynamics called ``group-affine observation systems". They were intially introduced for continuous-time dynamics \cite{barrau2014invariant,barrau2017linear} but are easily adapted to the discrete case \cite{barrau2017annual,barrau2017linear}. Here we briefly introduce them and recall some of their properties.

\begin{definition}
A discrete time system \eqref{eq:noisy_lie_system} with state $\chi_i$ at time $i$ being an element of a Lie group $G$ is said to be ``group-affine" if its dynamics/observation couple writes:
\begin{gather}
\forall i, k, \quad \begin{matrix}
\chi_{i+1} = f_i(\chi_i) \\
y_k = \chi_{t_k} d
\end{matrix} \label{eq:group_affine} \\
\text{where } \forall a,b \in G, f_i(ab) = f_i(a) f_i(Id)^{-1} f_i(b)
\end{gather}
\end{definition}
The dynamics part of this definition is called group-affine dynamics and has a very powerful property, that will lead to interesting results in the sequel. Note that a continuous-time counterpart of these dynamics also exists in the theory of control on Lie groups \cite{ayala1999linear}.

\begin{theorem}[from  \cite{barrau2014invariant}]
$f$ defines a group-affine dynamics if and only if $g^L(\chi) =: f(Id)^{-1} f(\chi)$ is a group automorphism, i.e., satisfies
\begin{equation}
 \forall a,b,\ g^L(ab) = g^L(a) g^L(b), \text{ and }g^L(a^{-1}) = g^L(a)^{-1}
\end{equation}
In that case, the Lie-group/Lie algebra morphism correspondance ensures the existence of a $q\times q$ matrix $G^L$ such that
\begin{equation}
\forall \xi, g^L(\exp(\xi)) = \exp(G^L \xi).
\label{eq:dg}
\end{equation}
\label{thm:group_affine}
\end{theorem}

\subsection{Noisy system}
For the remainder of this paper we will focus on the following dynamics, observation and noise models :
\begin{equation}
\chi_{i+1} = f_i(\chi_i) \exp(w_i) \qquad y_k = \chi_{t_k} d + v_k
\label{eq:noisy_lie_system}
\end{equation}
where $w_i$ and $v_k$ are white noise similar to those of \eqref{eq:noisy_linear_system},   $d$ is a given vector of $\RR^q$, and $f_i$ possesses the group-affine property \eqref{eq:group_affine} for all $i$.

In this model, the measurements are left-equivariant, a fact which according to \cite{barrau2014invariant}  prompts the use of the left multiplication based uncertainty model \eqref{eq:left}.  Therefore, given a current guess $\hat{\chi}$, the cost function will be linearized, noting $\xi$ the searched parameter, using
\begin{equation}
\chi = \hat{\chi} \exp(\xi).
\label{eq:right_error}
\end{equation}

\subsection{Factors definition}
\label{sec:factors}
Under the Gaussian and conditional independence assumptions, defining the factors boils down to isolating the noise in \eqref{eq:noisy_lie_system}.  We have $\exp(w_i)  = f_i(\chi_i)^{-1} \chi_{i+1}$. Let  $ \hat{\chi}_{i }, \hat{\chi}_{i+1}$ be some guesses, and let us write the true ${\chi}_{i },{\chi}_{i+1}$ using the Lie-group based error model \eqref{eq:right_error}. This yields 
\begin{align}
\exp(w_i) &= f_i(\chi_i)^{-1} \chi_{i+1}  \\
&=  f_i(\hat{\chi}_i \exp(\xi_i))^{-1} \hat{\chi}_{i+1} \exp(\xi_{i+1})
\end{align} Using \eqref{eq:group_affine}, and then \eqref{eq:dg} this is equal to
\begin{align*}
  &= (f_i(\hat{\chi}_i )f_i(Id)^{-1}f_i(\exp(\xi_i)))^{-1}  \hat{\chi}_{i+1} \exp(\xi_{i+1})\\
  &= (f_i(\hat{\chi}_i)g_i^L(\exp(  \xi_i)))^{-1}  \hat{\chi}_{i+1} \exp(\xi_{i+1})\\
  &= \exp(-G_i^L \xi_i) f_i(\hat{\chi}_i)^{-1} \hat{\chi}_{i+1} \exp(\xi_{i+1})
\end{align*} 
Let $a_i = \log(f_i(\hat{\chi}_i)^{-1} \hat{\chi}_{i+1})$ be the estimation error for the guess. Then, taking the logarithm of both hand sides and using the BCH formula, only keeping the first order terms in $\xi$ \emph{and} $a_i$ leads to
\begin{equation}
w_i \approx a_i - G_i^L \xi_i + \xi_{i+1}
\label{eq:propagation_factor}
\end{equation}
The major advantage (and very remarkable feature) of this parametrization is the fact that, if the dynamics is accurate enough the jacobians \emph{do not} depend on the current estimate since $G_i^L$ is independent of $\hat \chi$ from Theorem \ref{thm:group_affine}. This was also derived in the particular case of $SE(3) \times \RR^n$ in \cite{kobilarov2015dynamic}.

To get the measurement factors we propose to consider the following quantities and their linearization\begin{align}\label{quantities}
\hat{\chi}_{t_k}^{-1} v_k &= \hat{\chi}_{t_k}^{-1} y_k - \exp(\xi_{t_k}) d \\
 						&\approx c_k - H\xi_{t_k}
\label{eq:observation_factor}
\end{align} where $c_k = \hat{\chi}_{t_k}^{-1} y_k$, $H$ is the matrix defined by  $H\xi= \xi^\wedge d$, given by a Taylor expansion of the matrix exponential in \eqref{expmap}.

Finally, the factor associated to the prior must also be assessed. Indeed, the noise on the prior now writes
\begin{equation}
\exp(z_0) = \bar{\chi}_0^{-1} \hat{\chi}_0 \exp(\xi_0)
\label{eq:prior_factor}
\end{equation}
Contrarily to the propagation factor, here the first-order terms of the BCH formula cannot be reduced to the sum of its arguments, as the term $\log(\bar{\chi}_0^{-1} \hat{\chi}_0)$ will tend to grow over the iterations, sometimes to a point where the jacobian of the Lie group \cite{barfoot2017robotics} must be taken into account. We thus have in general
\begin{equation}
z_0 \approx p_0 + J_0 \xi_0 \Leftrightarrow J_0^{-1} z_0 \approx J_0^{-1} p_0 + \xi_0 = p_0 + \xi_0
\label{eq:initial_factor}
\end{equation}
where $p_0 = \log(\bar{\chi}_0^{-1} \hat{\chi}_0)$, and $J_0$ is defined as $BCH(p_0, \xi) = p_0 + J_0 \xi + o(\|\xi\|^2)$, which implies $J_0 p_0 = p_0$.

Therefore, the linear least-squares problem to be solved at each iteration writes
\begin{align}
(\xi_i)_{i\leq n} = &\argmin_{(\xi_i)_{i \leq n}} 
\|p_0 + \xi_0\|_{J_0^{-1} P_0 J_0^{-T}}^2  \nonumber \\
&+ \sum_i \|a_i - G^L_i \xi_i + \xi_{i+1}\|_{Q_i}^2 
+ \sum_k \|c_k + H \xi_{t_k}\|_{\hat{N}_k}^2
\label{eq:lie_group_lq}
\end{align}
where $\hat{N}_k = \hat{\chi}_{t_k}^{-1} N_k \hat{\chi}_{t_k}^{-T}$ is the covariance of $\hat{\chi}_{t_k}^{-1} v_k$, see \eqref{eq:observation_factor}.

\subsection{Final algorithm and benefits of the proposed approach}
\label{sec:benefits}
Injecting \eqref{eq:lie_group_lq} into the smoothing framework, we get Algorithm \ref{alg:smoothing} for the left multiplication based  parametrization \eqref{eq:right_error} (as opposed to $\chi = \exp(\xi) \hat{\chi}$). The main advantage of this framework are the following facts : $(i)$ The current estimates do not appear in the definition of the propagation and observation jacobians, in \eqref{eq:propagation_factor} and \eqref{eq:observation_factor} respectively (matrices $G^L$ and $H$), although the latter comes at the expense of a modified covariance for the measurement factors. $(ii)$ However, this dependency disappears if the measurement covariance $N_k$ is such that $N_k=\chi N_k \chi^T$ for all $\chi$. Although this condition covariance can seem restrictive at first, it often boils down to an isotropy assumption, as will be the case for the model of Section \ref{sec:exp}. This leaves only the prior covariance of \eqref{eq:prior_factor} to be a function of the estimate. $(iii)$ If the initial guess is  good, this can also be harmlessly relaxed by approximating $J_0 \approx Id$. If conditions $(i) - (iii)$ are met, the information matrix associated to \eqref{eq:lie_group_lq} can be considered independent from the current estimate.

\begin{algorithm}[t]
	\KwIn{$(\bar{\chi})_{1\leq i \leq n}, \mathbf{P}_0, (f_i)_i, (Q_i)_i, (y_k)_k, (N_k)_k$\;}
    \SetKwBlock{Initialization}{Initialization}{end}
    \Initialization{
    \nl Set $\hat{\chi}_i^0 = \bar{\chi}_i,\ i = 1, \cdots, n$
    }
	\SetKwBlock{Optimization}{Until convergence do}{end}
	\Optimization{
		\nl Linearize around $(\hat{\chi}_i^k)_i$ according to \eqref{eq:lie_group_lq} \;
        \nl Solve for $(\xi_i)_i$ \;
        \nl Update : $\hat{\chi}_i^{k+1} := \hat{\chi}_i \exp(\xi_i),\ i = 1, \cdots, n$
	}
    \nl Set $\chi_i^* = \hat{\chi}_i^k,\ i = 1, \cdots, n$ \;
	\KwOut{$(\chi_i)_{i \leq n}^*$\;}
	\caption{Smoothing for a group-affine system with a left multiplication based parametrization \eqref{eq:right_error}}
    \label{alg:smoothing}
\end{algorithm}

\section{Application to mobile robot localization}
\label{sec:exp}

\subsection{Considered problem: robot localization}
\label{sec:2d_nav}

To evaluate the performances of the developed approach, tests were conducted for a wheeled robot localization problem, using the standard non-linear equations of the 2D differential drive car modeling the position $x_i \in \RR^2$ and heading $\theta_i \in \RR$ of the robot. The odometer velocity is integrated between two time steps to give a position shift $u_i \in \RR^2$ and the angular shift $\omega_i \in \RR$ is measured through (differential) odometry and/or gyroscopes. The discrete noisy model writes:
\begin{align}
\theta_{i+1} &= \theta_i + \omega_i + w_i^\omega \nonumber \\
x_{i+1} &= x_i + R(\theta_i) (u_i + w_i^x)
\label{eq:discrete_dynamics}
\end{align}
where $w_i^\omega$ is the angular measurement error, $w_i^x$ contains both the odometry and transversal shift errors, and $R(\theta)\in SO(2)$ denotes the planar rotation of angle $\theta$. The vehicle also gets noisy  position measurements (through e.g., GNSS) with standard deviation $\sigma$ of the form
\begin{equation}
y_k = x_{t_k} + v_k,\qquad v_k \sim \mathcal{N} (\mathbf{0}, 
N_k = \sigma^2 I_2)
\label{eq:observations}
\end{equation}
The state (and the increments) can be embedded in the matrix Lie group $SE(2)$, which is detailed in Appendix \ref{app:se2}, using the homogeneous matrix representation $$\chi_i=:\begin{pmatrix}R(\theta_i)&x_i\\0_{1\times 2}&1\end{pmatrix},~ U_i=:\begin{pmatrix}R(\omega_i)& u_i \\0_{1\times 2}&1\end{pmatrix}$$  
and letting $\mathbf{w_i} = \begin{pmatrix} w_i^\omega & R(\omega_i)^T w_i^x \end{pmatrix}$ the stacked noise vector on the increments, we have using a first order expansion of the exponential map of $SE(2)$ (the increment noises are small if the time step is small)
$$\exp(\mathbf{w_i} )\approx \begin{pmatrix}R(w_i^\omega)&R(\omega_i)^T w_i^x \\0_{1\times 2}&1\end{pmatrix}.
$$
Thus, the dynamics and the observations respectively rewrite in the desired form \eqref{eq:noisy_linear_system}:
\begin{align}
\chi_{i+1} &= \chi_i U_i \exp(\mathbf{w_i}) = f_i(\chi_i) \exp(\mathbf{w_i})\\
y_k &= \chi_{t_k} \begin{pmatrix} 0_{2\times 1} \\ 1 \end{pmatrix} + \begin{pmatrix} v_k \\ 0 \end{pmatrix}
\end{align}

This system is group-affine and its measurement covariance is rotation-invariant (i.e., isotropic), therefore $(i)$ and $(ii)$ of \ref{sec:benefits} hold. Moreover, we have, with the notations of Section \ref{sec:smoothing_group} and according to Section \ref{sec:lie_groups},
\begin{equation}
g_i^L(\chi) = \Psi_{U_i^{-1}}(\chi)
\end{equation}

\subsection{Compared smoothing frameworks}
\label{sec:jacobians}
In Sections \ref{sec:simu} and \ref{sec:setting}, the proposed smoothing method is compared with two non-invariant parametrizations, but which account for the non linear structure of the state space, from respectively \cite{grisetti2010tutorial} and \cite{forster2016preintegration}, and a standard linear parametrization. The two former were developed for applications such as pose-SLAM or VIO, without directly considering navigation with absolute measurements. They were reimplemented in a batch setting, so that the focus was put exclusively on the parametrization. Writing the propagation linearized factor in the general form $a_i + F_i^i \xi_i + F_i^{i+1} \xi_{i+1}$, the following jacobians were used, see \cite{grisetti2010tutorial,forster2016preintegration} for the respective residuals' definitions :
\begin{center}
\begin{tabular}{c|c|c}
	&	$F_i^i$		&	$F_i^{i+1}$	 \\
	\hline
IS (ours)	&	$-Ad_{U_n^{-1}}$	&	$Id$	\\
Lin	&	$\begin{bsmallmatrix} -I_2 & -\hat{R}_i u_i \\ & -1 \end{bsmallmatrix}$ 	&	$Id$	\\
\cite{grisetti2010tutorial}	&	$\begin{bsmallmatrix} -\Omega_i^T\hat{R}_i^T 	&	J \Omega_i^T \hat{R}_i^T (\hat{x}_{i+1} - \hat{x}_i) \\ & -1 \end{bsmallmatrix}$	&	$\begin{bsmallmatrix} \Omega_i^T\hat{R}_i^T & \\ & 1 \end{bsmallmatrix}$ \\
\cite{forster2016preintegration}	&	$\begin{bsmallmatrix} I_2 & J \hat{R}_i^T (\hat{x}_{i+1} - \hat{x}_i)\\ & 1 \end{bsmallmatrix}$	&	$-\begin{bsmallmatrix} \hat{R}_i^T\hat{R}_{i+1}& \\ & 1 \end{bsmallmatrix}$
\end{tabular}
\end{center}

The observation factors are constructed using the following jacobians $H_k$, errors $c_k$ and covariances :

\begin{center}
\begin{tabular}{c|c|c|c}
	
	&	$H_k$		&	$c_k$	&	covariance  \\
	\hline
IS (ours)	&	$\begin{bsmallmatrix} I_2 & 0 \end{bsmallmatrix}$	&	$\hat{R}_{t_k}^T(y_k - \hat{x}_{t_k})$	&	$\hat{R}_{t_k}^T N_k \hat{R}_{t_k}$ \\
Lin, \cite{grisetti2010tutorial}	&	$\begin{bsmallmatrix} I_2 & 0 \end{bsmallmatrix}$ 	&	$y_k - \hat{x}_{t_k}$	&	$N_k$\\
\cite{forster2016preintegration}	&	$\begin{bsmallmatrix} \hat{R}_{t_k} & 0 \end{bsmallmatrix}$ 	&	$y_k - \hat{x}_{t_k}$	&	$N_k$
\end{tabular}
\end{center}


\subsection{Simulation results}
\label{sec:simu}
The four smoothing methods of the previous section were compared. The vehicle is initialized at the true position but with a wrong heading of $-3\pi /4$, and an initial covariance of $diag(0.0025, 0.0025, (-3\pi /4)^2)$ . The robot moves along a line at $7 m/s$, where the odometry is  polluted by a white noise of covariance $(0.1 (m/s)^2, 0.01 (rad/s)^2)$, and acquired at a rate of $10 Hz$. Position measurements of covariance $0.01 m^2$ are received every five steps. In this batch setting, $(iii)$ of \ref{sec:benefits} does not hold for the invariant framework.

\begin{figure}[thpb]
       \centering
       \includegraphics[width=.49\textwidth]{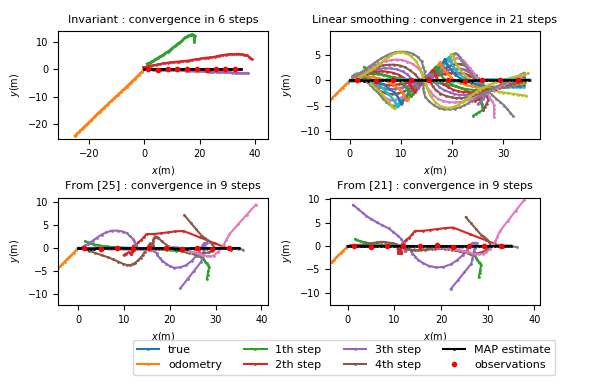}
       \caption{Comparison of the iterations of the smoothing algorithms based on four parametrizations : invariant (top-left), linear (top-right), from \cite{grisetti2010tutorial} (bottom-left), from \cite{forster2016preintegration} (bottom-right). Thanks to its being linearized independently from the estimate, the invariant version converges faster and in a more ``sensible" fashion.}
       \label{fig:batch_optim}
    \end{figure}

The iterations of the four optimization schemes are displayed on Figure \ref{fig:batch_optim}. As can be seen, the invariant parametrization converges faster than all the other ones, and much more smoothly. This is important for a sliding window smoothing setting, as restricting the number of solver iterations at each step will have a weaker impact on the estimation.

\subsection{Experimental setting}
\label{sec:setting}
\begin{figure}[thpb]
       \centering
       \includegraphics[width=.30\textwidth, scale=0.2]{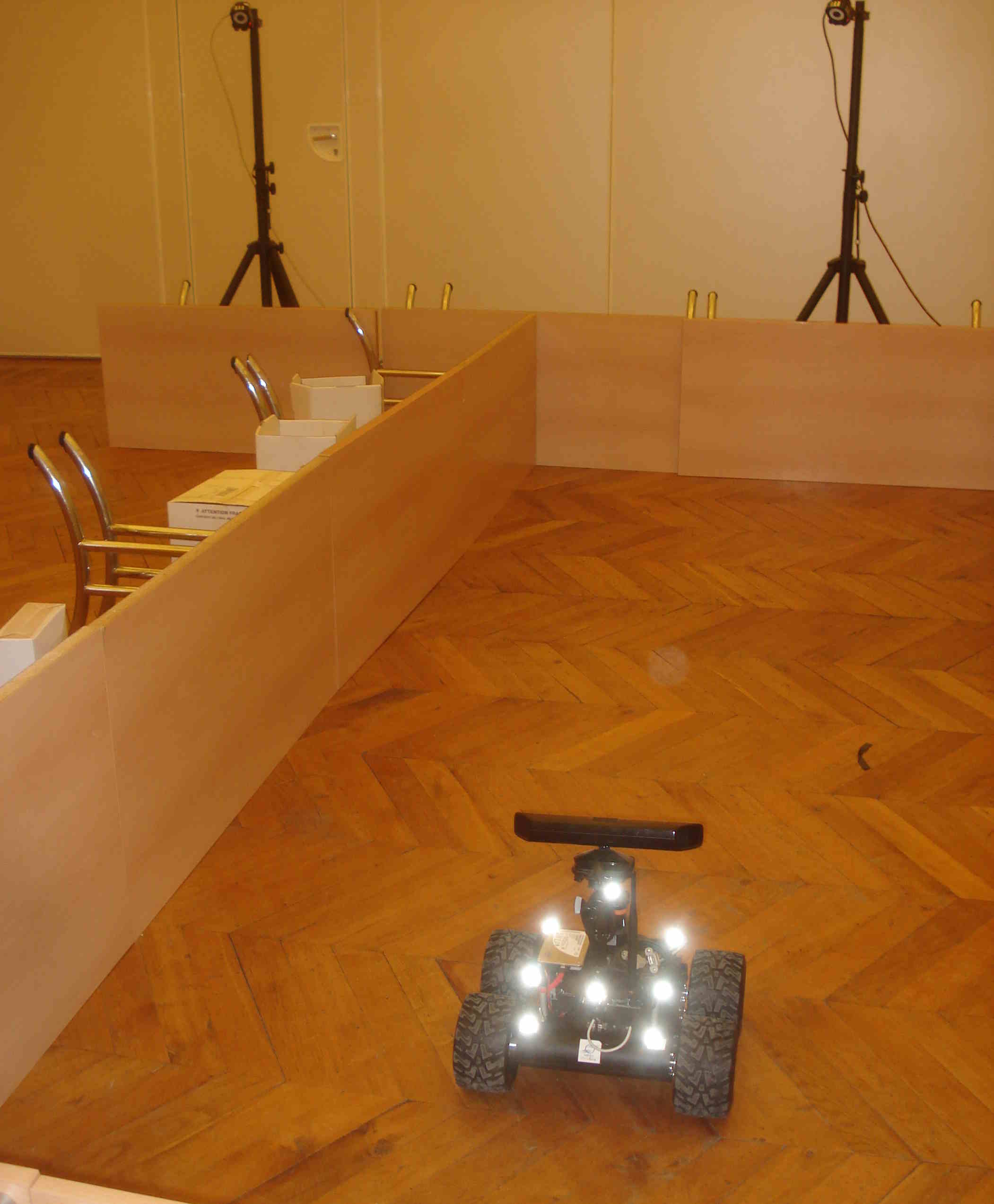}
       \caption{Wifibot robot in its testing arena, surrounded by Optitrack cameras}
       \label{fig:robot}
    \end{figure}

\begin{figure}[thpb]
       \centering
       \includegraphics[width=.45\textwidth]{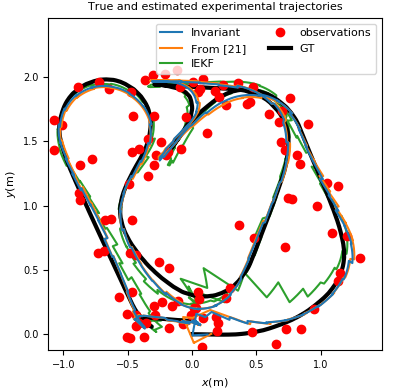}
       \caption{Trajectories of the wifibot robot, as captured by the OptiTrack system and  estimated by three of the studied methods : invariant smoothing, smoothing based on \cite{forster2016preintegration}, both using a sliding window of size 5, and the IEKF, with artificial noisy measurements obtained by adding simulated moderate noise -  $\sigma = 0.01 (m/s)^2$  - to the ground truth  position, showed by the red dots. }
       \label{fig:trajectory}
    \end{figure}

We compare the proposed method to various other methods based on data obtained in an experiment conducted at the Centre for Robotics, MINES ParisTech. A small wheeled robot, called Wifibot and photographed in Figure \ref{fig:robot}, is equipped with independent odometers on the left and right wheels. We made it follow an arbitrary trajectory for 80 seconds. The OptiTrack motion capture system, a set of seven highly precise cameras, provides the ground truth with sub-millimeter precision at a rate of 120 Hz. This choice allows us to directly compare our results with other recent algorithms based on the invariant framework, namely the Left-UKF-LG, which had already been tested on these data \cite{brossard2017unscented}, the Invariant EKF, and both non-invariant smoothing methods from Section \ref{sec:simu}. To keep the computational complexity acceptable, the smoothing is done in a sliding window, the oldest state being marginalized out when a new one is added. This choice makes $(iii)$ hold in this case, leading to an information matrix fully independent from the estimate. The effect of the window's size is also studied here. The raw odometer inputs were provided to each method. Artificial position measurements are delivered by adding a Gaussian noise to the ground truth, at a rate of $1.35 Hz$. For each setting, 100 Monte-Carlo simulations are run, each algorithm being initialized identically at each run. The trajectory followed by the robot, along with results of the various algorithms for one of the runs, are displayed in Figure \ref{fig:trajectory}. Three series of experiments were conducted, to study the evolution of the error with respect to the measurement covariance, the window size, and the initial error respectively. In the first case, the window size was fixed to 5, and the measurement covariance ranged from $10^{-1}$ to $10^{-5} m^2$, with noise turned on and off on the initial state. When on, $(\bar{x}_0, \bar{\theta}_0) \sim \mathcal{N}( (0,0,0), \text{diag}(1/8, 1/8, (\pi/4)^2)$. When off, it was put at $\bar{x}_0 = (1/4, 1/4)$ and $\bar{\theta}_0 = \pi/4$. In the second one, the measurement covariance was fixed to $10^{-1} m^2$, while the window size ranged from 5 to 13. For the last one, the window size was set to 5, the measurement covariance to $10^{-5} m^2$. The initial estimate was put at $\bar{x}_0 = (1/4, 1/4)$ and $\bar{\theta}_0$ spanned $]-\pi, \pi[$ with only one Gauss-Newton per iteration, then it was fixed at $\bar{\theta}_0 = 9\pi / 10$ with one and seven GN iterations. Figures \ref{fig:RMSE_vs_cov} and \ref{fig:RMSE_vs_window} show the averaged Root Mean Square Error (RMSE) of the position and heading for the two first experiments. Figure \ref{fig:RMSE_vs_angle} shows the RMSE of the heading for the last one.

\begin{rem}
Average computation times of the full estimate \emph{with one GN iteration per step}, on a laptop with Intel i5-5300 2.3 GHz CPU, are given in the following table.

\begin{center}
\begin{tabular}{|c|cccc|}
\hline
	&	Invariant	&	From \cite{forster2016preintegration}	&	From \cite{grisetti2010tutorial} 	&	IEKF \\
    \hline
 Time (s) & 0.82 &  0.81	&	0.83	&	0.62
\label{tab:computation_time}
\end{tabular}
\end{center}

It appears that the invariant smoothing, in a batch setting, does not induce extra computational load using $\exp$ and $\log$. Further gain could be expected with more GN iterations per step, and it could also prove more efficient in an incremental setting by taking advantage of the independence of the information matrix from the linearization points to prefactor the information matrix \cite{kaess2012iSAM2}.
\end{rem}

\subsection{Experimental results}
\label{sec:results}
\begin{figure}[thpb]
       \centering
       \includegraphics[width=.44\textwidth]{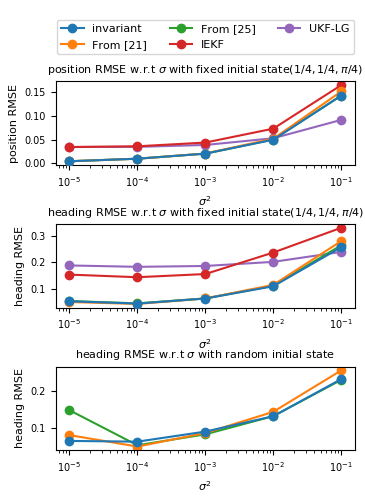}
       \caption{Top and middle : average position and heading RMSE for the five estimation methods w.r.t $\sigma$, with fixed initial state. Down : heading RMSE w.r.t. $\sigma$, with random initial state. The window size was set to 5 for the smoothing methods.}
       \label{fig:RMSE_vs_cov}
    \end{figure}

Figure \ref{fig:RMSE_vs_cov} displays the evolution of the error with respect to the measurement noise covariance, when the initial state is fixed and when it is not. 
The first observation to be made is that, for a noise up to moderate ($10^{-2} m^²$), smoothing methods in general outperform the filtering ones, especially in terms of heading. This was expected, as the former explicitly contains the constraints between successive poses. If the initial state is fixed at 
$(1/4, 1/4, \pi/4)$, all smoothing approaches achieve almost identical results validating, in this case, the assumption made when deriving the propagation factor. However, invariant smoothing appeared to be the most robust to noise on the initial state, especially for low measurement covariance. This is further investigated in the third experiment.

\begin{figure}[thpb]
       \centering
       \includegraphics[width=.42\textwidth]{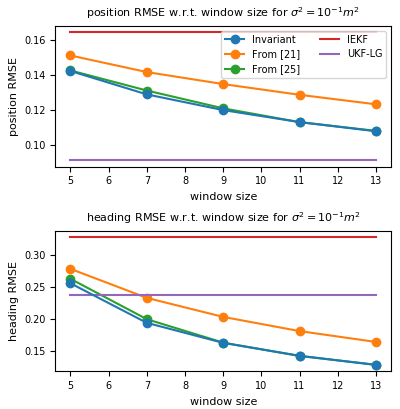}
       \caption{Average position and heading RMSE for the smoothing methods w.r.t. the window size, for $\sigma^2 = 10^{-1} m^2$. The RMSE of the filtering methods are shown as a comparison.}
       \label{fig:RMSE_vs_window}
    \end{figure}

The impact of the window size, illustrated in Figure \ref{fig:RMSE_vs_window}, seems clear: increasing the size of the window is beneficial to all smoothing methods for strong noise. Still, none of them beats the UKF-LG in terms of position, highlighting its robustness. However, for lighter noise, the impact was negligible, as the odometry's uncertainty was quickly reached.

\begin{figure}[thpb]
       \centering
       \includegraphics[width=.45\textwidth]{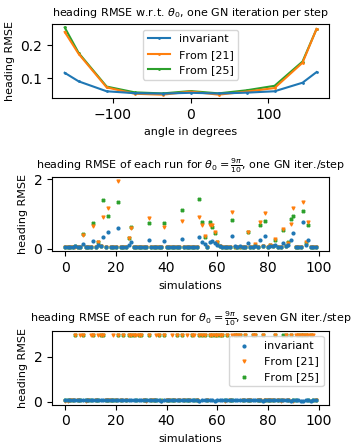}
       \caption{Heading RMSE for the smoothing methods w.r.t the initial angle $\bar{\theta}_0$ for $\sigma^2 = 10^{-5} m^2$ and a window size of 5. Top : $\bar{\theta}_0$ spans $]-\pi, \pi]$ with one GN iteration per step. Middle : RMSE for each run with $\bar{\theta}_0 = 9 \pi / 10$ with one GN iteration. Down :  RMSE for each run with $\bar{\theta}_0 = 9 \pi / 10$ with seven GN iterations.}
       \label{fig:RMSE_vs_angle}
    \end{figure}

Finally, the invariant smoothing proved to be the more robust to initial heading errors such that $|\bar{\theta}_0| > \pi/2$, as shown by Figure \ref{fig:RMSE_vs_angle}, top, while both non invariant methods behaved identically. The two other graphs explain this difference : it appears that non-invariant smoothing methods fall into a local minimum where the estimate is completely turned around. Indeed, when increasing the number of GN iterations, the non-invariants experiments clearly form two clusters, one with the expected heading RMSE, and one with a RMSE close to $\pi$. Note that it is not a problem of the non-invariant methods, as they were devised for problems relative to the original position, i.e. without such an influence of the initial error. Moreover, this confirms the results of Section \ref{sec:simu}, as the invariant smoothing reaches the optimal RMSE quicker than the non-invariant ones, in the right cases.
\subsection{Discussion}
\label{sec:discussion}
For this experiment, the proposed method managed to combine the advantages of the smoothing approach and of the invariant framework. Indeed, on the first hand it provided better estimates than the filtering methods thanks to the smoothing paradigm. On the other hand, owing to the mathematical invariances leveraged by the algorithm, their computation are based on the measured odometry inputs, and not the estimated ones. This reduces the need for relinearization, while ensuring more robustness to initial heading error, especially when non-invariant methods exhibited local minima with shifted heading estimates. However, this also means that the method will be dependent of the odometry's uncertainties.

\section{Conclusion}
This paper presented the extension of  the invariant Kalman filtering framework (see e.g., \cite{barrau2017annual} for an overview)  to smoothing. This resulted in Jacobians being independent from the current estimate, thus ensuring sound behavior of the estimate. To validate the results, this was applied to a robot localization problem, in simulations, and using experimental data. On the one hand, faster convergence to the MAP estimate than the other smoothing approaches was achieved in simulation. On the other hand, the invariant smoothing provided estimates similar to that of other smoothing methods, while being as fast to compute, and better than invariant filtering methods. It also appeared more robust to errors on the initial state, avoiding local minima of the non-invariant methods. Future work will include extending the framework to relative measurements of the state.

\section{Acknowledgements}
The authors would like to thank Martin BARCZYK and Tony NO\"EL for their precious help with the experiments.
This work is supported by the company Safran through the CIFRE convention 2016/1444.

\appendices
\section{Operators of the Special Euclidean 2D Group}
\label{app:se2}
The Special Euclidean 2D $SE(2)$ group represents rigid transformations in 2D space defined by their heading $\theta$ and position $x$. The exponential, logarithm and adjoint operators on $SE(2)$ are defined for $\chi \in SE(2)$ and $\xi \in \mathfrak{se}(2)$ as 

\begin{equation}
\exp\left(\xi\right) = \setlength\arraycolsep{1pt}\begin{bmatrix} R\left(\xi^3\right) & \mathbf{V}_{\xi^3} \begin{bmatrix}
\xi^1 \\ \xi^2
\end{bmatrix}
\\
\mathbf{0} & 1
\end{bmatrix},
\setlength\arraycolsep{1pt}
\log \left(\chi\right) = \begin{bmatrix} \mathbf{V}^{-1}_{\theta} x \\
\theta\end{bmatrix},
\label{eq:exp_log_SE2}
\end{equation}
\begin{equation}
\text{where } \mathbf{V}_{\alpha} = \frac{1}{\alpha} \begin{bmatrix}
\sin \left( \alpha \right) & -1+\cos \left( \alpha \right) \\
1-\cos \left( \alpha \right) & \sin \left( \alpha \right)
\end{bmatrix}
\end{equation}
\begin{equation}
Ad_{\chi} = \begin{bmatrix}
R(\theta) & -J x \\
\mathbf{0} & 1
\end{bmatrix} \mbox{, where }
J = \begin{bmatrix} 0 & -1 \\ 1 & 0 \end{bmatrix}
\end{equation}

\begin{spacing}{0.85}

\bibliography{biblio}

\begin{thebibliography}{10}

\bibitem{absil2008optim}
Pierre-Antoine Absil, Robert Mahony, and Rodolphe Sepulchre.
\newblock {\em Optimization Algorithms on Matrix Manifolds}.
\newblock Princeton University Press, Princeton, NJ, 2008.

\bibitem{ayala1999linear}
Victor Ayala and Juan Tirao.
\newblock Linear control systems on lie groups and controllability.
\newblock In {\em Proceedings of symposia in pure mathematics}, volume~64,
  pages 47--64. AMERICAN MATHEMATICAL SOCIETY, 1999.

\bibitem{barfoot2017robotics}
Timothy~D. Barfoot.
\newblock {\em State Estimation for Robotics}.
\newblock Cambridge University Press, 2017.

\bibitem{barfoot2014associating}
Timothy~D. Barfoot and Paul.~T. Furgale.
\newblock {Associating Uncertainty with Three-Dimensional Poses for Use in
  Estimation Problems}.
\newblock {\em IEEE Transactions on Robotics}, 30(3):679--693, 2014.

\bibitem{barrau2015SLAM}
Axel Barrau and Silv{\`e}re Bonnabel.
\newblock An {EKF-SLAM} algorithm with consistency properties.
\newblock {\em CoRR}, abs/1510.06263, 2015.

\bibitem{barrau2015intrinsic}
Axel Barrau and Silv{\`e}re Bonnabel.
\newblock Intrinsic filtering on lie groups with applications to attitude
  estimation.
\newblock {\em IEEE Transactions on Automatic Control}, 60(2):436 -- 449, 2015.

\bibitem{barrau2014invariant}
Axel Barrau and Silv{\`e}re Bonnabel.
\newblock The invariant extended kalman filter as a stable observer.
\newblock {\em IEEE Transactions on Automatic Control}, 62(4):1797--1812, April
  2017.

\bibitem{barrau2017linear}
Axel Barrau and Silvere Bonnabel.
\newblock Linear observation systems on groups (i).
\newblock 2017.

\bibitem{barrau2017annual}
Axel Barrau and Silv{\`e}re Bonnabel.
\newblock Invariant kalman filtering.
\newblock {\em Annual Review of Control, Robotics, and Autonomous Systems},
  1(1):null, 2018.

\bibitem{bonnabel2008symmetry}
Silv{\`e}re Bonnabel, Philippe Martin, and Pierre Rouchon.
\newblock {Symmetry-Preserving Observers}.
\newblock {\em {IEEE Transactions on Automatic Control}}, 53(11):2514--2526,
  December 2008.

\bibitem{boumal2013rotations}
Nicolas Boumal, Amit Singer, and Pierre-Antoine Absil.
\newblock Robust estimation of rotations from relative measurements by maximum
  likelihood.
\newblock In {\em 52nd IEEE Conference on Decision and Control}, pages
  1156--1161, Dec 2013.

\bibitem{bourmaud2016online}
Guillaume Bourmaud.
\newblock {\em Online Variational Bayesian Motion Averaging}, pages 126--142.
\newblock Springer International Publishing, Cham, 2016.

\bibitem{bourmaud2015concentrated}
Guillaume Bourmaud, R{\'e}mi M{\'e}gret, Marc Arnaudon, and Audrey Giremus.
\newblock {Continuous-Discrete Extended {K}alman Filter on Matrix Lie Groups
  Using Concentrated Gaussian Distributions}.
\newblock {\em Journal of Mathematical Imaging and Vision}, 51(1):209--228,
  2015.

\bibitem{bourmaud2016iterated}
Guillaume Bourmaud, R{\'e}mi M{\'e}gret, Audrey Giremus, and Yannick
  Berthoumieu.
\newblock From intrinsic optimization to iterated extended kalman filtering on
  lie groups.
\newblock {\em J. Math. Imaging Vis.}, 55(3):284--303, July 2016.

\bibitem{brossard2017unscented}
Martin Brossard, Silv{\`e}re Bonnabel, and J.P. Condomines.
\newblock Unscented kalman filtering on lie groups.
\newblock In {\em 2017 IEEE/RSJ International Conference on Intelligent Robots
  and Systems (IROS)}, pages 2485--2491, Sept 2017.

\bibitem{chirikjian2009stochastic1}
Gregory~S. Chirikjian.
\newblock {\em Stochastic Models, Information Theory, and Lie Groups, Volume 1:
  Classical Results and Geometric Methods}.
\newblock Applied and numerical harmonic analysis. Birkh\"auser, 2009.

\bibitem{chirikjian2011stochastic2}
Gregory~S Chirikjian.
\newblock {\em Stochastic Models, Information Theory, and Lie Groups, Volume 2:
  Analytic Methods and Modern Applications}.
\newblock Springer Science \& Business Media, 2011.

\bibitem{dellaert2006square}
Frank Dellaert and Michael Kaess.
\newblock Square root sam: Simultaneous localization and mapping via square
  root information smoothing.
\newblock {\em The International Journal of Robotics Research},
  25(12):1181--1203, 2006.

\bibitem{diemer2015controller}
S{\'e}ebastien Diemer and Silv{\`e}re Bonnabel.
\newblock {An Invariant Linear Quadratic Gaussian Controller for a Simplified
  Car}.
\newblock In {\em Robotics and Automation (ICRA), 2015 IEEE International
  Conference on}, pages 448--453. IEEE, 2015.

\bibitem{farrell2008aided}
Jay Farrell.
\newblock {\em Aided Navigation: GPS with High Rate Sensors}.
\newblock McGraw-Hill, Inc., New York, NY, USA, 1 edition, 2008.

\bibitem{forster2016preintegration}
Christian Forster, Luca Carlone, Frank Dellaert, and Davide Scaramuzza.
\newblock On-manifold preintegration for real-time visual--inertial odometry.
\newblock {\em IEEE Transactions on Robotics}, 33(1):1--21, Feb 2017.

\bibitem{forster2017SVO2}
Christian Forster, Zichao Zhang, Michael Gassner, Manuel Werlberger, and Davide
  Scaramuzza.
\newblock Svo: Semidirect visual odometry for monocular and multicamera
  systems.
\newblock {\em IEEE Transactions on Robotics}, 33(2):249--265, April 2017.

\bibitem{gallier2016diffgeo}
J.~Gallier.
\newblock Notes on differential geometry and lie groups, 2016.
\newblock University of Pennsylvannia.

\bibitem{grisetti2010hierarchical}
Giorgio Grisetti, Raimer K{\"u}mmerle, Cyrill Stachniss, Udo Frese, and
  Christoph Hertzberg.
\newblock Hierarchical optimization on manifolds for online 2d and 3d mapping.
\newblock In {\em 2010 IEEE International Conference on Robotics and
  Automation}, pages 273--278, May 2010.

\bibitem{grisetti2010tutorial}
Giorgio Grisetti, Ramier K{\"u}mmerle, Cyrill Stachniss, and Wolfram Burgard.
\newblock A tutorial on graph-based slam.
\newblock {\em IEEE Intelligent Transportation Systems Magazine}, 2(4):31--43,
  winter 2010.

\bibitem{hertzberg2013integrating}
Christoph Hertzberg, Ren{\'e} Wagner, Udo Frese, and Schr{\"o}der Lutz.
\newblock {Integrating Generic Sensor Fusion Algorithms with Sound State
  Representations Through Rncapsulation of Manifolds}.
\newblock {\em Information Fusion}, 14(1):57--77, 2013.

\bibitem{indelman2012factor}
Vadim Indelman, Stephen Williams, Michael Kaess, and Frank Dellaert.
\newblock Factor graph based incremental smoothing in inertial navigation
  systems.
\newblock In {\em Information Fusion (FUSION), 2012 15th International
  Conference on}, pages 2154--2161. IEEE, 2012.

\bibitem{kaess2012iSAM2}
Michael Kaess, Hordur Johannsson, Richard Roberts, Viorela Ila, John~J.
  Leonard, and Frank Dellaert.
\newblock {iSAM2}: Incremental smoothing and mapping using the {B}ayes tree.
\newblock {\em The International Journal of Robotics Research}, 31:217--236,
  2012.

\bibitem{kobilarov2015dynamic}
Martin Kobilarov, Duy-Nguyen Ta, and Franck Dellaert.
\newblock Differential dynamic programming for optimal estimation.
\newblock In {\em 2015 IEEE International Conference on Robotics and Automation
  (ICRA)}, pages 863--869, May 2015.

\bibitem{kuemmerle2011g2o}
Rainer K{\"{u}}mmerle, Giorgio Grisetti, Hauke Strasdat, Kurt Konolige, and
  Wolfram Burgard.
\newblock g2o: A general framework for graph optimization.
\newblock In {\em 2011 IEEE International Conference on Robotics and
  Automation}, pages 3607--3613, May 2011.

\bibitem{mahony2017geometric}
Robert~E. Mahony and Tarek Hamel.
\newblock A geometric nonlinear observer for simultaneous localisation and
  mapping.
\newblock In {\em 56th {IEEE} Annual Conference on Decision and Control, {CDC}
  2017, Melbourne, Australia, December 12-15, 2017}, pages 2408--2415, 2017.

\bibitem{martinec2007robust}
Daniel Martinec and Tom{\'a}{\v s} Pajdla.
\newblock Robust rotation and translation estimation in multiview
  reconstruction.
\newblock In {\em 2007 IEEE Conference on Computer Vision and Pattern
  Recognition}, pages 1--8, June 2007.

\bibitem{rosen2017SE-sync}
David~M. Rosen, Luca Carlone, Afonso~S. Bandeira, and John~J. Leonard.
\newblock Se-sync: {A} certifiably correct algorithm for synchronization over
  the special euclidean group.
\newblock {\em CoRR}, abs/1612.07386, 2016.

\bibitem{rosen2015relaxation}
David~M. Rosen, Charles DuHadway, and John~J. Leonard.
\newblock A convex relaxation for approximate global optimization in
  simultaneous localization and mapping.
\newblock In {\em 2015 IEEE International Conference on Robotics and Automation
  (ICRA)}, pages 5822--5829, May 2015.

\bibitem{rosen2014RISE}
David~M. Rosen, Michael Kaess, and John~J. Leonard.
\newblock Rise: An incremental trust-region method for robust online sparse
  least-squares estimation.
\newblock 30(5):1091--1108, October 2014.

\bibitem{wolfe2011bayesian}
Kevin~C. Wolfe, Michael Mashner, and Gregory~S. Chirikjian.
\newblock {Bayesian Fusion on Lie groups}.
\newblock {\em Journal of Algebraic Statistics}, 2(1):75--97, 2011.

\bibitem{zhao2014differential}
Sheng Zhao, Yiming Chen, Haiyu Zhang, and Jay~A. Farrell.
\newblock Differential gps aided inertial navigation: a contemplative realtime
  approach.
\newblock {\em IFAC Proceedings Volumes}, 47(3):8959 -- 8964, 2014.
\newblock 19th IFAC World Congress.

\end{thebibliography}
\bibliographystyle{plain}

\end{spacing}

\end{document}